# Extensions of Morse-Smale Regression with Application to Actuarial Science

By Colleen M. Farrelly[1]

[1]Independent Researcher, cfarrelly@med.miami.edu

**Abstract:**

The problem of subgroups is ubiquitous in scientific research (ex. disease heterogeneity, spatial distributions in ecology…), and piecewise regression is one way to deal with this phenomenon. Morse-Smale regression offers a way to partition the regression function based on level sets of a defined function and that function's basins of attraction. This topologically-based piecewise regression algorithm has shown promise in its initial applications, but the current implementation in the literature has been limited to elastic net and generalized linear regression. It is possible that nonparametric methods, such as random forest or conditional inference trees, may provide better prediction and insight through modeling interaction terms and other nonlinear relationships between predictors and a given outcome.

This study explores the use of several machine learning algorithms within a Morse-Smale piecewise regression framework, including boosted regression with linear baselearners, interaction-term homotopy-based LASSO, conditional inference trees, random forest, and a wide neural network framework called extreme learning machines. Simulations on Tweedie regression problems with varying Tweedie parameter and dispersion suggest that many machine learning approaches to Morse-Smale piecewise regression improve the original algorithm's performance, particularly for outcomes with lower dispersion and linear or a mix of linear and nonlinear predictor relationships. On a real actuarial problem, several of these new algorithms perform as good as or better than the original Morse-Smale regression algorithm, and most provide information on the nature of predictor relationships within each partition to provide insight into differences between dataset partitions.

**Introduction**

Piecewise regression has offered a good solution for problems involving subpopulations and spatially-based data (McZgee & Carleton, 1970); applications include engineering (Boys et al., 2016; Ryan, Porth, & Troendle, 2002), ecology (Toms & Lesperance, 2003), criminology (Stalans & Seng, 2007), actuarial science (Apte et al., 1999; De John & Heller, 2008; DeWeerdt & Dercon, 2006; Farrelly, 2017; Panjer & Willmot, 1992), and medicine (Chassin, Pitts, & Prost, 2002; Compston et al., 1980; Farrelly et al., 2013; Kehagia, Barker, & Robbins, 2010; McClellan & King, 2010; Tomoda et al., 2016). One recent development involving the partitioning of a dataset based on Morse-Smale clustering from the field of topology has shown some promise in analyzing and visualizing data (Gerber et al., 2013); however, it is largely limited to regression models involving only main effects terms. Multivariate methods that can capture interaction terms, such as random forest, have shown promise on many regression and classification problems (Fernandez-Delgado et al., 2014), suggesting that these methods may be able to improve Morse-Smale regression performance and predictive capability.

Several recent machine learning extensions of generalized linear regression exist and have shown promise in regression problems, including neural network models, tree-based models, boosted ensembles, and bagged ensembles (Fernandez-Delgado et al., 2014). Ensemble methods build robust models through combining weak baselearner models, either through a series of bootstrapped models or through iterative averaging of models grown on the error of previous models (Breiman, 2001; Friedman, 2001); random forest and boosted regression both show good performance on a wide variety of problems. Despite this success, some single tree models also show fairly good performance, though different data samples can produce different trees. Conditional inference trees are fairly robust, as they employ statistical testing methods at each split (Hothon, Hornik, & Pascucci, 2006).

Neural networks, particularly extreme learning machines (ELMs) with proven universal approximation properties, have also performed well on a variety of regression problems (Huang, Zhu, & Siew, 2004). ELMs randomly map data space to a series of hidden nodes; this allows for the use of least squares methods (based on Moore-Penrose inverses) rather than the usual backpropagation to fit the network, increasing computational speed dramatically (Huang, Zhu, & Siew, 2004).

This paper explores the use of Morse-Smale regression with several machine learning methods to compute the regression piece, comparing performance of the piecewise regressions with the performance of the original algorithm utilizing main effects elastic net models. Simulations include a wide variety of Tweedie-distributed outcomes, covering many types of regression problems; significant improvements are found with several of the proposed algorithms, particularly for linear or mixed models with low to moderate dispersion. These algorithms are then applied to an insurance dataset, predicting 1977 claims payouts for a Swedish 3rd party motor insurance company. Most Morse-Smale-based regression methods perform relatively well, particularly ensemble-based methods and the original algorithm.

**Methods**

**1) Tweedie regression overview**

Tweedie distributions belong to the exponential family, where links are defined to generalize linear regression to non-normal outcomes, which are called generalized linear models (Tweedie, 1984); Tweedie distributions focus on the variance function and its relationship to the mean. Many common exponential family distributions converge to Tweedie distributions and can be formulated through Tweedie distributions, including normal distributions, Poisson distributions, gamma distributions, and compound Poisson-gamma distributions (Tweedie, 1984). Formally, the mean and variance of a Tweedie distribution are given by:

$$E(Y) = \mu$$

$$Var(Y) = \varphi \mu^{\xi}$$

where $\varphi$ is the dispersion parameter, and $\xi$ is the Tweedie parameter (or shape parameter).

## 2) Morse-Smale regression overview

Topological approaches to data analysis have had many recent success, including clustering (Nicolau, Levine, & Carlsson, 2011), data comparison (Chen, Genovese, & Wasserman, 2017), and data visualization/summarization (Wasserman, 2016; Zomorodian, 2012; Zomorodian & Carlsson, 2005). This field, called topological data analysis, has been applied to problems in medicine (Nicolau, Levine, & Carlsson, 2011; Alagappan et al., 2016), actuarial science (Farrelly, 2017), psychometrics (Farrelly et al., 2017), medical imaging (Lee et al., 2012), time series mining (Pereira & de Mello, 2015), and network analysis (Lee et al., 2012). Morse-Smale regression, a supervised learning algorithm in topological data analysis, is a piecewise elastic net model, where pieces are defined though decomposing the model space defined by KNN-based local neighborhoods though differential topology (Gerber et al., 2013). Morse functions, a special type of continuous function analogous to the height function used in topography (Forman, 2002; Mischaikow & Nanda, 2013), are used to identify nondegenerate critical points of a data manifold and their basins of attraction (technically intersections of descending and ascending manifolds that form the Morse-Smale complex). These basins of attraction—which are based on the Hessian matrix—partition the data manifold into discrete pieces that share local minima and maxima (Chen et al., 2017); regression models are then fit to these pieces (Gerber et al., 2013).

An intuitive analogy of how the algorithm explores a low-dimensional space to identify local minima and maxima, as well as basins of attraction that partition that space, is a soccer player dribbling a soccer ball (Figure 1). When he reaches a local maximum and kicks the ball, it will roll towards a local minimum, defining a gradient descent flow in the field. Continuing this strategy, he eventually explores the entire field, defining the descending manifolds; dribbling up the hills in a similar manner defines the ascending manifolds. Taken together, these pieces for the Morse-Smale complex, partitioning the field into pieces of similar gradients and local optima.

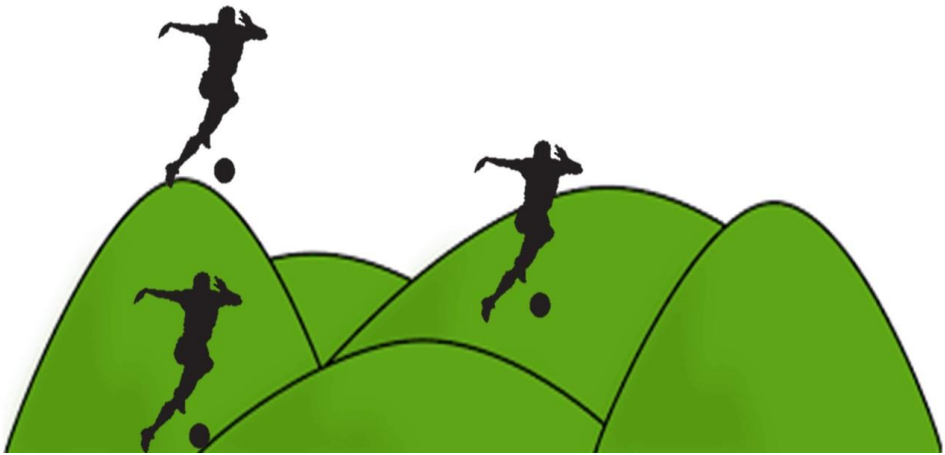

*Figure 1: Morse-Smale Complex Analogy*

Of models yielding interpretable insight into predictor-outcome relationships, Morse-Smale regression performs well compared to state-of-the-art machine learning algorithms (Farrelly, 2017).

## 3) Proposed algorithms overview

### 3.1) Random forest

The random forest algorithm is a well-known and widely-used bagging algorithm in which trees are grown from bootstrap samples, generally drawing 1/3 of the predictor variables in one bootstrap sample, and then aggregated the collection of fully-grown trees into a final model (Breiman, 2001). Given a training dataset, D, with an outcome, Y, and a collection of predictors, X, over n observations:

$$D_n = \{(X_i, Y_i)\}_{i=1}^n$$

the random forests model with M trees yields an estimate of the outcome class or measurement, $T_m(X)$, modeling Y based on the collection of weights, $W_{im}$, taken across all trees and observations from bootstrap samples of predictor variables, X:

$$F(X) = \frac{1}{M} \sum_{m=1}^{M} T_m(X) = \frac{1}{M} \sum_{m=1}^{M} \sum_{i=1}^{n} W_{im}(X) Y_i$$

with each individual tree in the forest contributing its model, specified by

$$T_m(X) = \sum_{i=1}^{n} W_{im}(X) Y_i$$

Variables across all trees in this model are then ranked by importance, which can be obtained through permutation testing of the $j^{th}$ predictor entered into the model. The $j^{th}$ predictor is randomly permuted within the dataset, and the change in prediction error is calculated from the left-out portion of data within bootstrap iterations (Breiman, 2001).

Intuitively, this is like compiling a full book report on a long novel, such as *East of Eden,* by randomly assigning students in a literature course chapters of the book to read and summarize. With enough chapters assigned per student and enough students in the course, the set of summaries will capture most chapters and provide a good overview of the book (though likely a bit disjointed in this case).

### 3.2) Boosted regression

Gradient boosting methods model an outcome with an iteratively updated link function, $F_m(x)$, composed of a weighted sum of base learner functions, $h_i(x)$, and their weights, $\gamma_i$,

$$F(x) = \sum_{i=1}^{M} \gamma_i h_i(x) + constant$$

updated in a greedy fashion with respect to minimizing the expected value of a certain loss function such that model components with large errors in a given iteration are given preferential weighting in the subsequent iteration to correct the model error (Friedman, 2001). An analogy for this type of algorithm is deducing the main parts of a puzzle from key pieces that are iteratively added to form a clearer and clearer image of the full scene.

Because this problem of adaptively adding model base learners is computationally difficult, steepest descent methods based on gradient calculation are used to update the boosted model (Friedman, 2001), similar to a climber deciding upon the quickest route of down the mountain and represented as:

$$F_m(x) = F_{m-1}(x) - \gamma_m \sum_{i=1}^{n} \nabla(L(\gamma_i, F_{m-1}(x_i)))$$

where the weight given to baselearners at a particular iteration is given as:

$$\gamma_m = \arg\min_{\gamma} ||y - X\beta|| \sum_{i=1}^{m} L\left(\gamma_i, F_{m-1}(x_i) - \gamma \frac{\partial L(\gamma_i, F_{m-1}(x_i))}{\partial f(x_i)}\right)$$

in which $L(y_i, F_{m-1}(x_i))$ is a pre-specified loss function, such as $\ell_1$ loss or Huber loss. Optimizing weights in the model through a line search gives:

$$\gamma_m = \arg\min_{\gamma} \sum_{i=1}^{m} L(\gamma_i, F_{m-1}(x_i) + \gamma h_m(x_i))$$

where the update is based upon minimization of the loss function through gradient descent. The result of this process creates a stable, strong final model after m iterations (which is either user-specified by a minimum change in loss function or a set maximum number of iterations):

$$F_m(x) = F_{m-1}(x) + \gamma_m h_m(x)$$

As such, boosting offers a stable and flexible solution to the logistic regression problem and can handle situations in which predictors outnumber observations or predictors do not have strictly linear relation to outcome (Friedman, 2001).

### 3.3) Homotopy LASSO and Penalized Regression Models

Penalized methods, such as elastic net and LASSO, extend generalized linear regression to samples in which predictors may outnumber observations by imposing penalties to the least squares estimator:

$$\min_{\beta} ||y - X\beta|| \text{ s.t. } ||\beta||_1 = \sum_{j=1}^{p} |\beta_j| \leq t$$

where p is the number of predictors and t is the regularization parameter, such that predictors with an estimated coefficient less than t are set to 0 (Osborne, Presnell, & Turlach, 2000). Geometrically, this can be thought of as a cowboy standing at the origin casting his lasso of a given length to capture and bag for removal any predictors that fall within the radius of his cast.

The elastic net method used in the original Morse-Smale regression algorithm extends this form of $\ell_1$ norm penalty to $\ell_1$ and $\ell_2$ penalties (Gerber et al., 2013), such that the equation becomes:

$$\min_{\beta} ||y - X\beta|| \text{ s.t. } J \leq t$$

where

$$J(\boldsymbol{\beta}) = \alpha ||\boldsymbol{\beta}||^2 + (1 - \alpha) \, ||\boldsymbol{\beta}||_1$$

Thus, elastic net can create a sparse model (from the $\ell_1$ penalty) that also has a unique minimum (from the $\ell_2$ penalty creating a convex penalty search area). Essentially, elastic net combines a LASSO shrinkage with a ridge regression model for robust performance with shrinkage of terms; this allows the algorithm to handle situations where predictors outnumber observations, as well as situations where predictors are correlated with each other or exhibit deviations from linearity. This is analogous to cleaning up the area around the cowboy such that obstacles (posts, trees…) are removed from his field of vision for easier roping of variables that get too close to his position at the origin.

An extant LASSO extension, homotopy LASSO (Osborne, Presnell, & Turlach, 2000), includes a topologically-based search of model coefficients. This method leverages properties of the Lagrange multiplier associated with the $\ell_1$ penalty, which yields an ordinary differential equation (ODE) solution that is piecewise linear at all points where the constraint is differentiable. ODEs with piecewise linearity have a known solution involving the deformation of paths from a known equation's solution to a target equation solution that is unknown or difficult to solve. This is known as the homotopy method (Osborne, Presnell, & Turlach).

Homotopy is an intrinsic property related to the equivalence of paths between points, in which paths are equivalent if they can be continuously deformed to one another (Figure 2). In Figure 2, all paths on the sphere can be continuously deformed to each other; on the torus (the donut-shaped space), some paths are hindered by the hole and cannot be continuously deformed to each other. Thus, homotopy search for regression parameters can search for easy predictor paths and then deform them according to the data, avoiding geometric pitfalls that may exist in the parameter/data spaces that can trap other algorithms. By analogy, this method is like a blindfolded person trying to reach a target from which he is connected by a rope through a field of obstacles; without a rope to guide him around the obstacles, he may be temporarily or permanently unable to get around them to the final target.

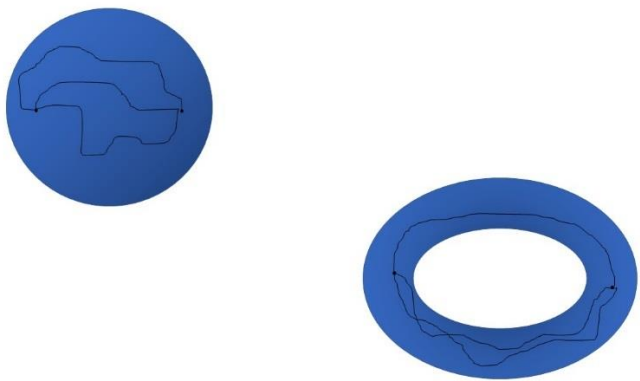

*Figure 2: Homotopy Path Example*

## 3.4) Conditional Inference Tree

Trees are nonparametric models built by recursively splitting a dataset's space according to a rule that will minimize heterogeneity with each subdivision until a stopping criterion is met (Breiman, 2001). At each split, a set of predictor variables is assessed according to these criteria, and the variable producing the best split is chosen to maximize model fit (Breiman, 2001). Typically, this evaluation rule involves measuring the model prediction error or mean square error for each potential splitting variable. Regardless of split criteria measure chosen, this process stops either when a set of conditions is met (such as minimum node size or a threshold of splitting criterion change) or when a certain iteration number is reached (Breiman, 2001). Geometrically, this amounts to a recursive partitioning of hyperspace into regions with progressively more homogenous data points.

Conditional inference trees are based on statistically testing whether these partitions of hyperspace involve covariates that are independently distributed with respect to the outcome or if the covariates are conditionally distributed with respect to the outcome (Hothorn, Hornik, & Zeileis, 2006). This induces recursive partitioning of space into conditionally-dependent regions. Geometrically, this is equivalent to finding overlapping regions of space, which create nonempty sets intersections (Figure 3).

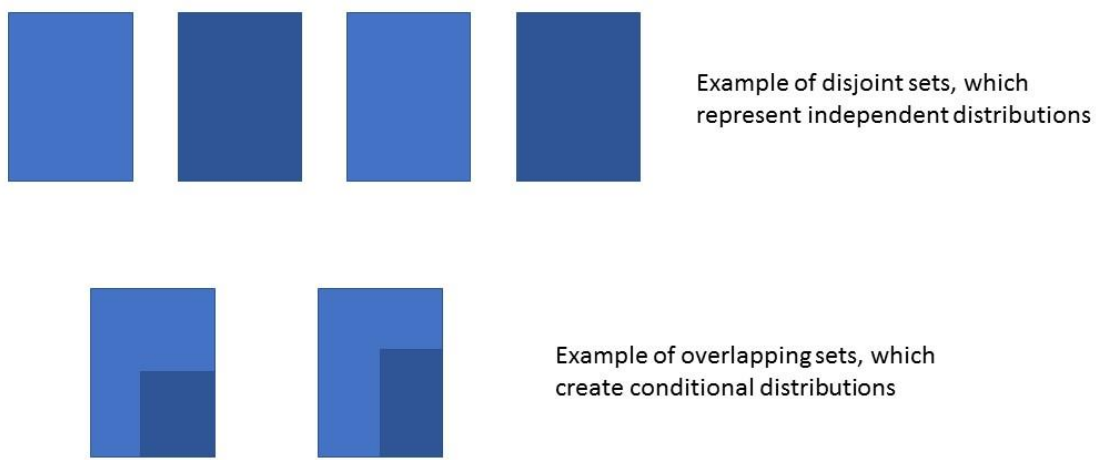

*Figure 3: Conditional and Independent Subspaces*

### 3.5) ELM

Neural networks rely on a series of topological mappings to process the data and seek to minimize error between those mappings given an outcome (Huang, Zhu, & Siew, 2004). Extreme learning machines (ELMs) employ a single, wide layer of hidden nodes with random mapping of features to hidden nodes. This random mapping allows for the use of least squares methods to optimize weights between the hidden layer and output (as opposed to costly algorithms like gradient descent techniques). Technically, ELMs seek a solution to:

$$\|H\boldsymbol{\beta}_{hat} - T\| = \min_{\beta}\|H\boldsymbol{\beta} - T\|$$

which yields a solutions of:

$$\boldsymbol{\beta}_{hat} = \boldsymbol{H}^\dagger \boldsymbol{T}$$

where $\boldsymbol{H}^\dagger$ is the Moore-Penrose inverse, $\boldsymbol{H}$ is the hidden node parameters, and $\boldsymbol{T}$ is the training outcome (Huang, Zhu, & Siew, 2004). This method has shown promise on regression problems (Fernandez-Delgado et al., 2014), and given its universal approximation property, extreme learning machines can model a wide variety of regression models.

### 3.6) Morse-Smale Extension Set-Up

The general approach to extending Morse-Smale regression to multivariate regression models involved 3 steps: partitioning the data via Morse-Smale complexes, fitting the multivariate models to each partition using the training dataset, and evaluating the performance on the test dataset (Figure 4). This set-up decreases computational cost, as the partitions are created upfront, but limits the applicability of this particular implementation for online or batch algorithms.

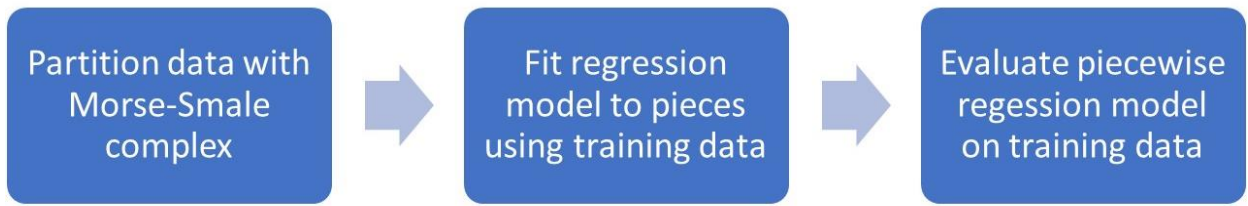

*Figure 4: Morse-Smale Regression Algorithm Construction and Testing Steps*

All code was written in R using the msr package for Morse-Smale complex computations and regression, ranger for random forest, mboost for boosted regression, partykit for conditional inference trees, lasso2 for homotopy LASSO, and elmNN for ELMs. Custom code used to create each hybrid algorithms is available upon request.

### 4) Simulation Set-Up

Simulations were used to test algorithm performance across a wide variety of problems with known truths. All simulations involved 4 true predictors and 11 noise variables and a Tweedie-distributed outcome. Sample size was set to 10000, as potentially small complex sizes complicates convergence of machine learning algorithms; at 10000, most complexes involved >150 individuals. Trials involved varying the nature of predictor to outcome relationship (all linear, all nonlinear, or a mix of linear and nonlinear), Tweedie parameter (1 for Poisson, 1.5 for compound Poisson-gamma, and 2 for gamma), and dispersion (1 for normal dispersion, 2 for moderate overdispersion, 4 for more overdispersion). Data was split 70/30 to create a training set and a test set upon which to evaluate algorithm performance. Each trial was run 10 times, and mean square error (MSE) was averaged across the 10 trials.

Algorithms tested included the original Morse-Smale regression algorithm with elastic net model (MSR), conditional inference tree (TR) and its Morse-Smale model (MSTR), random forest (RF) and its Morse-Smale model (MSRF), extreme learning machine (ELM) and its Morse-Smale model (MSELM), main-effect+interaction-term boosted regression (BR) and its Morse-Smale model (MSBR), and homotopy LASSO (LH) and its Morse-Smale model (MSLH).

**5) Swedish Motor Insurance Trial**

The Swedish 3rd party motor insurance dataset is an actuarial dataset from http://www.statsci.org/data/glm.html, which consists of predicting the value of 1977 payouts from insurance policies. Its 6 predictors include kilometers traveled a year, geographical zone, bonus, car model make, number of years insured, and total number of claims; this dataset contains 2182 observations. Again, data was split into train and test sets with a 70/30 split. Models were compared with mean model and original Morse-Smale regression through MSE.

**Results**

**1) Simulation**

In general, results suggest that using multivariate models with Morse-Smale piecewise regression framework outperform the original algorithm (Figures 5, 6, and 7). This is particularly true for trials involving linear or mixed predictor relationships and trials with lower dispersion. Some Morse-Smale algorithms outperformed their non-piecewise counterparts; other times, the non-piecewise models outperformed the Morse-Smale-based ones. However, piecewise models allow for comparison of predictor relationships within population subgroups and visualization of groups by predictors and outcome, which may be useful in data modeling projects.

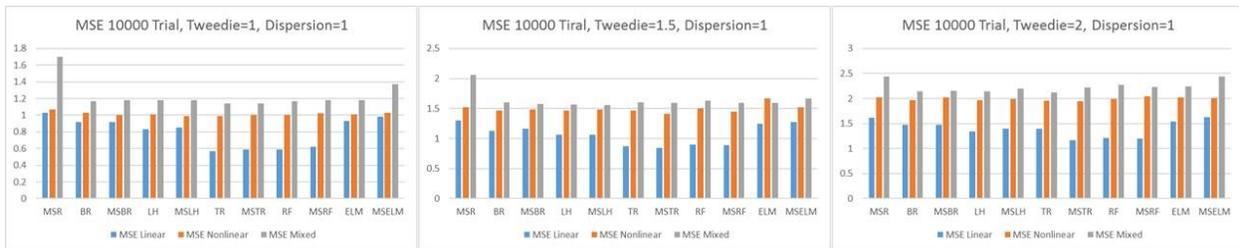

Figure 5: Dispersion=1 Trials

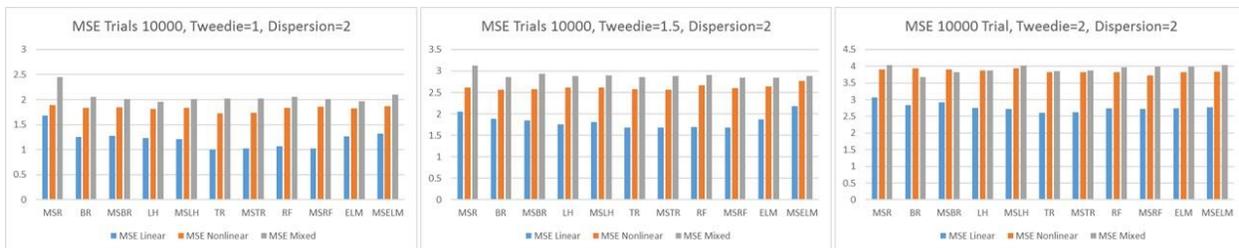

Figure 6: Dispersion=2 Trials

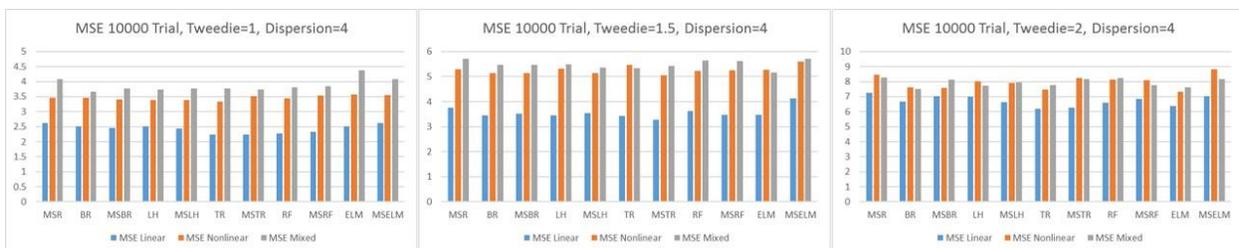

Figure 7: Dispersion=4 Trials

For instance, on a gamma regression trial with no overdispersion, partitions yielded groups with 569, 9278, and 153 individuals; random forest regressions yielded different variable importance scores and rankings. These differences are captured in both the importance graphs (Figure 8 upper) and the low-dimensional representation of the data in the associated Morse-Smale plot, where differences between variables in the data space represented on a series of axes (Figure 8 lower). The Morse-Smale plot highlights the very different multivariate composition of these groups with respect to the outcome of interest; the random forest plot summarizes this in a more typical chart of differences that might be more useful in the communication of results. This is a key advantage of piecewise models, particularly those based on Morse-Smale complexes, which come with visualization software components to explore group differences.

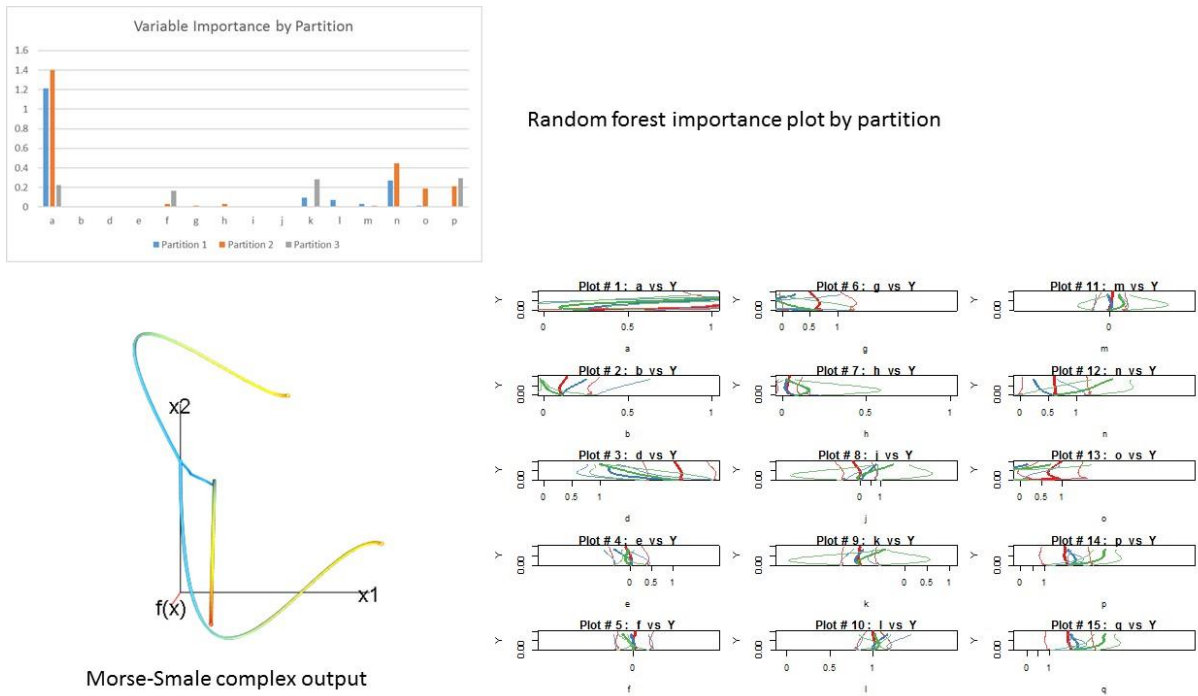

*Figure 8: Deep Dive of Dispersion=1, Tweedie=2 Simulation with Morse-Smale Random Forest Regression*

**2) Swedish Motor Insurance Trial**

Most models performed well on the Swedish Motor Insurance dataset, with Morse-Smale-based random forest achieving the best overall performance (Figure 9). This suggests that many machine learning methods can perform well on this dataset, particularly the ensemble methods and Morse-Smale-based methods. Dramatic improvements in the homotopy LASSO algorithm using the piecewise approach suggest that this approach may fix singularities or troublesome geometric characteristics that can trip up least squares algorithms.

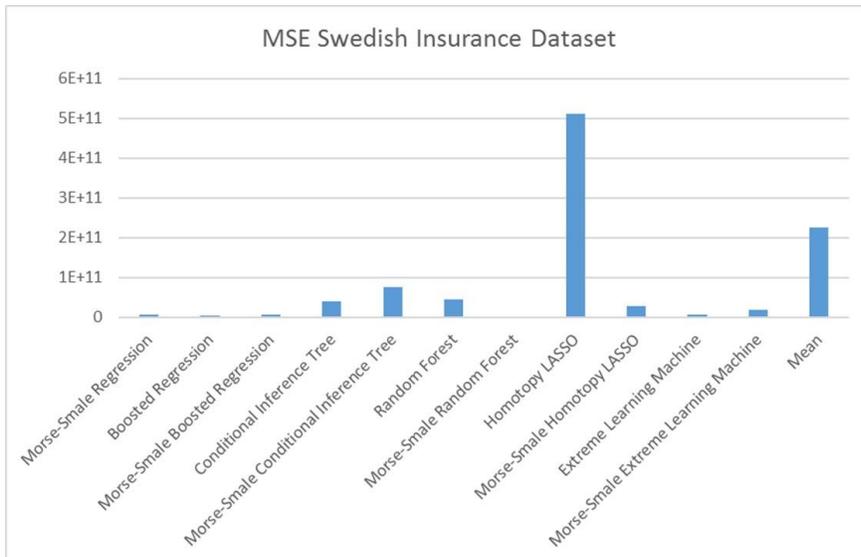

*Figure 9: Swedish Motor Insurance Results (MSE)*

Again, group differences emerge (Figure 10), and the dataset is split into a large group (N=2039) and 2 smaller groups (N=52, N=91). Morse-Smale plots show some differences among the groups with respect to kilometers, zone, bonus, and make; the random forest partition importance graph shows one group with relatively small importance on insured, with another group showing no effect for the kilometers or zone predictors. This gives a good overview of group differences and how multivariate importance can differ across subpopulations; piecewise regression, particularly those based on Morse-Smale complexes, allows for visual exploration of these differences while providing good predictive capabilities.

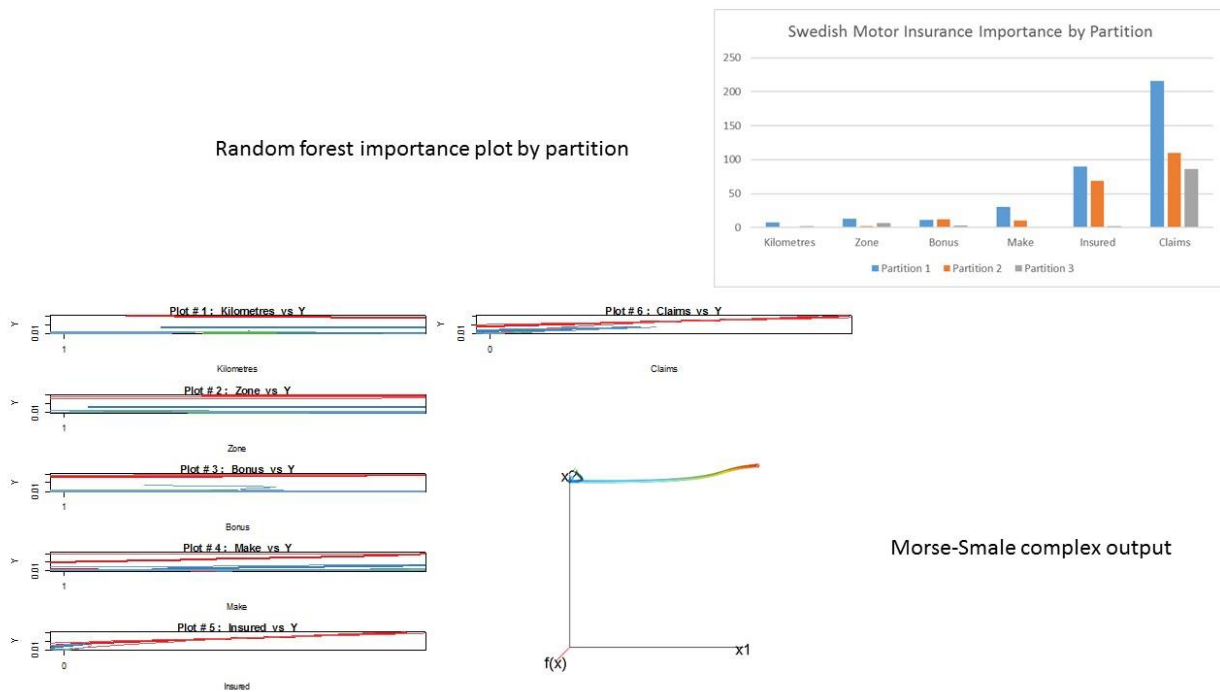

*Figure 10: Swedish Motor Insurance Results by Partition*

## Discussion

Morse-Smale regression as originally formulated can be applied to a wide variety of regression tasks (any involving continuous outcomes); extensions of this method to nonparametric regression algorithms within Morse-Smale partitions provide better prediction and multivariate insight into group differences. This is particularly true on problems with extant linear relationships within the dataset and low-to-moderate dispersion.

One advantage of these new algorithms is the ability to visualize results and draw inference on subpopulations. Morse-Smale partitions allow for the visualization of predictor relationships across groups, and saved regression models from each partition can provide visual diagrams and rankings of predictor importance for each partition. This can add important information about the nature of the predictor relationships and classification of subpopulations based on these relationships that may be useful in stratifying risk models or creating heuristic policy rules. While other methods perform better on Tweedie regression problems—such as KNN regression ensembles—they lack the inherent interpretability of relationships between predictors and the outcome of interest (Farrelly, 2017).

A limitation of this study is the reliance on Morse-Smale complexes in the partitioning of the piecewise models. Other methods of partitioning may result in better data splits, reducing overall error. In addition, the coding used to create these models does not allow for adaptive addition of samples, such that new data can be assigned into an extant partition. This precludes the use of this particular implementation in online applications or batch sets; however, it is possible to extend the R code developed to create this type based on the original Morse-Smale regression package.

Another limitation is the fixed sample size; while this examines behavior of the algorithms after convergence, it does not shed light on algorithm performance on small datasets. Future studies may want to examine under what conditions convergence is reached to guide application to smaller datasets. This is particularly important with small partitions, as sample sizes may be too small to obtain good prediction from certain methods like ELMs and other neural networks.

In all, this paper demonstrates the efficacy of multivariate models within piecewise regression frameworks, with particular emphasis on insurance data and actuarial applications.